\documentclass[cameraready]{Interspeech}

\title{DuplexCascade: Full-Duplex Speech-to-Speech Dialogue with VAD-Free Cascaded ASR–LLM–TTS Pipeline and Micro-Turn Optimization}

\author[affiliation={1,2}]{Jianing}{Yang}
\author[affiliation={1}]{Yusuke}{Fujita}
\author[affiliation={1}]{Yui}{Sudo}

\address{
    $^1$ SB Intuitions Corp., Japan, $^2$ The University of Tokyo, Japan
}

\email{baleyang@g.ecc.u-tokyo.ac.jp, yusuke.fujita@sbintuitions.co.jp, yui.sudo@sbintuitions.co.jp}

\keywords{Full-Duplex Dialogue; Speech-To-Speech Systems; Cascaded Architectures; Turn-Taking Control}

\usepackage{makecell}
\usepackage{multirow}
\usepackage{comment}
\usepackage{cite}

\begin{document}

\maketitle

\begin{abstract}
    Spoken dialog systems with cascaded ASR–LLM–TTS modules retain strong LLM’s intelligence, but VAD segmentation often forces half-duplex turns and brittle control. On the other hand, VAD-free end-to-end model supports full-duplex interaction but is hard to maintain conversational intelligence. In this paper, we present DuplexCascade, a VAD-free cascaded streaming pipeline for full-duplex speech-to-speech dialogue. Our key idea is to convert conventional utterance-wise long turns into chunk-wise micro-turn interactions, enabling rapid bidirectional exchange while preserving the strengths of a capable text LLM. To reliably coordinate turn-taking and response timing, we introduce a set of conversational special control tokens that steer the LLM’s behavior under streaming constraints. On Full-Duplex-Bench and VoiceBench, DuplexCascade delivers state-of-the-art full-duplex turn-taking and strong conversational intelligence among open-source speech-to-speech dialogue systems.

\end{abstract}

\section{Introduction}

\begin{figure*}[t]
    \centering
    \includegraphics[width=0.90\textwidth]{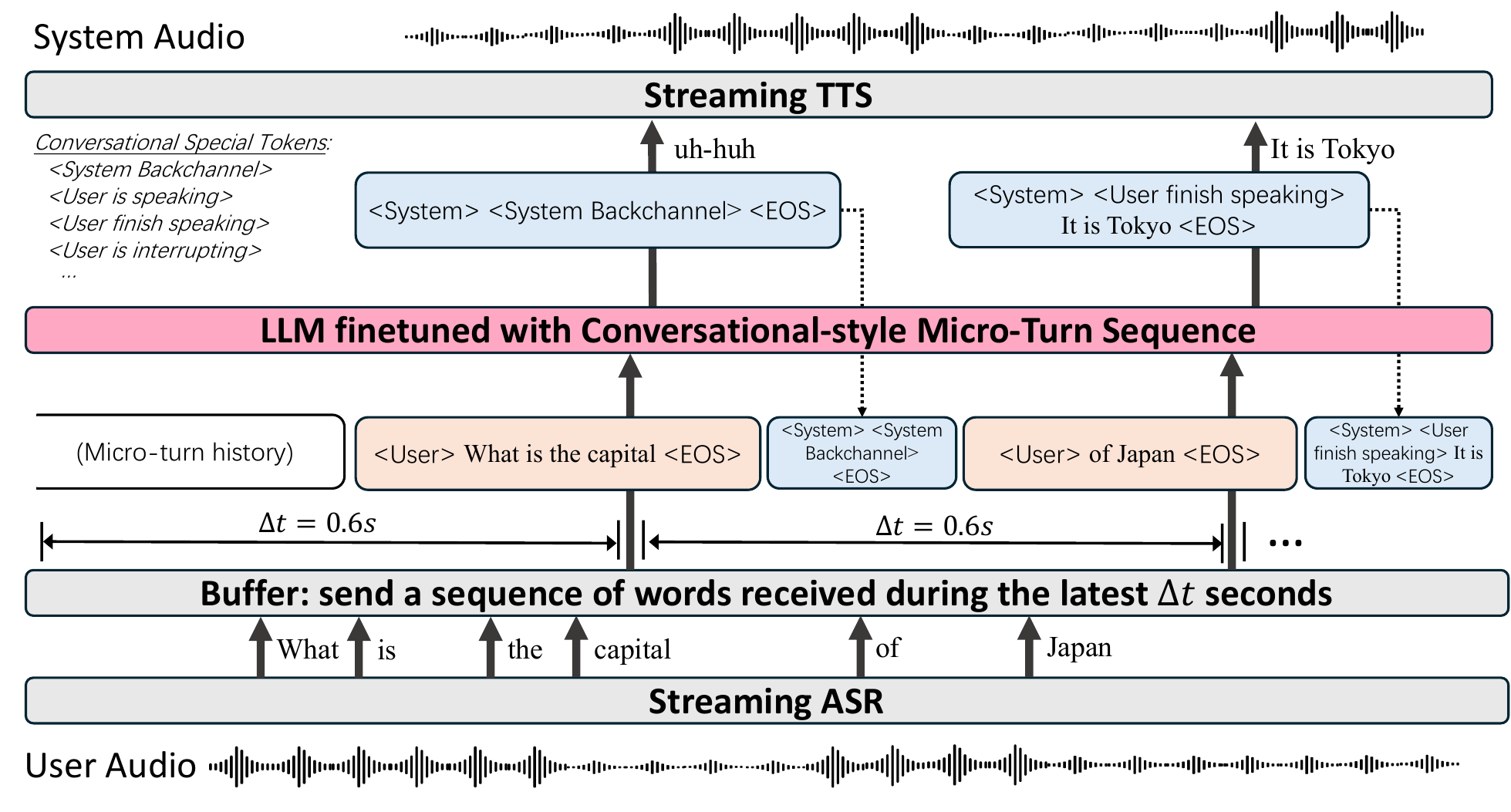}
    
    \caption{\normalfont Overview of DuplexCascade. User audio is transcribed by a streaming ASR and periodically flushed into text micro-turns ($\Delta t=0.6\,$s). The LLM consumes the dialogue history and the latest micro-turn to generate the next system micro-turn together with conversational special tokens (e.g., wait, respond, or backchannel). The generated text is then synthesized by a streaming TTS to produce system audio, enabling full-duplex interaction.}
    
    \label{fig:pipeline}
\end{figure*}

Recent advances in large language models (LLMs) have significantly improved the quality and usability of spoken dialogue systems, bringing speech-to-speech assistants closer to natural, low-friction interaction~\cite{wavchat2024}. In practice, most deployed systems adopt a cascaded design in which automatic speech recognition (ASR) transcribes user speech~\cite{shangguan20_interspeech}, an LLM performs dialogue reasoning in the text domain, and text-to-speech (TTS) synthesizes the system response~\cite{saeki2021incremental, torgashov2025voxtream}. This modular architecture remains attractive because it is straightforward to engineer and iterate on, and it inherits the strong instruction-following and reasoning capabilities of modern text LLMs.

Despite their strong conversational intelligence, cascaded systems typically depend on an external voice activity detector (VAD) to segment user speech into turns. This reliance often forces a half-duplex “listen-then-speak” interaction style and makes turn control brittle under pauses, overlaps, and noise, leading to unnatural turn-taking behaviors such as missing backchannels or failing to handle mid-utterance interruptions~\cite{freeze-omni, shannon2017eoq, liang2023dynamicendpoint, skantze2021turntaking}.

In contrast, recent end-to-end (E2E) speech--text models aim to remove explicit VAD-based segmentation and directly support concurrent listening and speaking, enabling more natural full-duplex turn control~\cite{kyutai2024moshi, roy2026personaplex, nguyen2023generative}. However, this VAD-free duplex capability often comes with a practical downside: compared with strong text LLMs, conversational intelligence can degrade due to the difficulty of jointly learning robust cross-modal representations and dialogue policies~\cite{chen2024voicebench}.


In this work, we propose DuplexCascade\footnote{\url{https://sbintuitions.github.io/DuplexCascadeDemo}}, a cascaded streaming pipeline that achieves full-duplex speech-to-speech dialogue while preserving the intelligence benefits of an LLM-centered design. The key idea is to transform conventional utterance-wise long turns into chunk-wise micro-turn interactions, enabling more natural full-duplex dialogue. We further introduce a carefully designed set of conversational special tokens to explicitly regulate LLM behaviors critical for duplex interaction, enabling VAD-free turn-taking control and making the overall policy more stable and controllable. With only 50k multi-turn text dialogues, we perform lightweight LoRA adaptation~\cite{hu2022lora} for 5k steps, achieving strong full-duplex performance on Full-Duplex-Bench~\cite{lin2025full}. Moreover, the text-only adaptation avoids cross-modal alignment issues, yielding strong conversational intelligence performace on VoiceBench~\cite{chen2024voicebench}.


\section{Related Works}

\subsection{VAD-Based Turn Taking Control}
Most recent speech-to-speech dialogue systems in practice follow a cascaded design with ASR, LLM, and TTS components, and rely on an external VAD to decide turn boundaries~\cite{shannon2017eoq,maas2018eou,ko2018vad}. However, VAD-based endpointing is fundamentally limited because it has difficulty leveraging conversational semantics and intent, which are often crucial for making precise turn-taking decisions~\cite{raux2008endpointing,liang2023dynamicendpoint}. As a result, cascaded systems often suffer from unstable turn-taking behaviors: they may interrupt the user when a pause is semantically meaningful, or remain silent when an immediate response or backchannel is expected~\cite{skantze2021turntaking,maier2017deepeot,castillolopez2025turntakingsurvey}. Recent cascade-style systems such as Freeze-Omni~\cite{freeze-omni} inherit this difficulty, and VAD errors frequently become the dominant failure mode for turn-taking quality.

\subsection{VAD-Free Time-Division-Multiplexing Approach Toward Full-Duplex Interaction}
Beyond VAD-driven cascades, recent work has explored making LLM-centered systems behave in a full-duplex manner by restructuring interaction into short, alternating segments without using external VAD~\cite{zhang2024beyond,veluri2024syncllm,zhang2025omniflatten}. MiniCPM-Duplex~\cite{zhang2024beyond} introduces the concept of Time-Division-Multiplexing encoding-decoding, which divides an interaction into time slices and processes each input slice immediately to produce the corresponding output slice. While this formulation enables pseudo full-duplex interaction after fine-tuning, it does not explicitly model or control turn-taking decisions; consequently, performance can degrade in realistic dialogue settings.


\section{DuplexCascade}
\begin{table*}[t]
\centering
\caption{\normalfont Full-Duplex-Bench results. The best-performing duplex dialogue model is shown in \textbf{bold}, and the second best is \underline{underlined}. $^{\S}$ means the results are infered from offical checkpoints. $^{\dagger}$ means the results directly obtained
from the paper.}
\resizebox{\textwidth}{!}{
\begin{tabular}{l cc ccc cc ccc c}
\toprule
\textbf{Dimension}
& \multicolumn{2}{c}{\textbf{Pause Handling}}
& \multicolumn{3}{c}{\textbf{Backchannel}}
& \multicolumn{2}{c}{\textbf{Smooth Turn Taking}}
& \multicolumn{3}{c}{\textbf{User Interruption}}
& \multirow{2}{*}{\makecell[c]{\textbf{Averaged}\\\textbf{Turn-Taking}\\\textbf{Accuracy}}} \\
\cmidrule(lr){2-3}\cmidrule(lr){4-6}\cmidrule(lr){7-8}\cmidrule(lr){9-11}
\textbf{Data Metric}
& \begin{tabular}[c]{@{}c@{}}\textbf{Synthetic}\\ \textbf{TOR ($\downarrow$)}\end{tabular}
& \begin{tabular}[c]{@{}c@{}}\textbf{Candor}\\ \textbf{TOR ($\downarrow$)}\end{tabular}
& \textbf{TOR ($\downarrow$)}
& \begin{tabular}[c]{@{}c@{}}\textbf{ICC}\\ \textbf{Freq ($\uparrow$)}\end{tabular}
& \textbf{JSD ($\downarrow$)}
& \begin{tabular}[c]{@{}c@{}}\textbf{Candor}\\ \textbf{TOR ($\uparrow$)}\end{tabular}
& \textbf{Latency ($\downarrow$)}
& \textbf{TOR ($\uparrow$)}
& \begin{tabular}[c]{@{}c@{}}\textbf{Synthetic}\\ \textbf{GPT-4o($\uparrow$)}\end{tabular}
& \textbf{Latency ($\downarrow$)} \\
\midrule
dGSLM$^{\dagger}$ & 0.934 & 0.935 & 0.691 & 0.015 & 0.934 & \textbf{0.975} & 0.352 & 0.917 & 0.201 & 2.531 & 0.466 \\
Moshi$^{\dagger}$ & 0.985 & 0.980 & 1.000 & 0.001 & 0.957 & \underline{0.941} & \underline{0.265} & \textbf{1.000} & 0.765 & \underline{0.257} & 0.395\\
Freeze-Omni$^{\dagger}$ & 0.642 & 0.481 & 0.636 & 0.001 & 0.997 & 0.336 & 0.953 & 0.867 & 3.615 & 1.409 & 0.489 \\
PersonaPlex$^{\dagger}$ & 0.358 & \underline{0.431} & \underline{0.273} & \textbf{0.042} & \textbf{0.662} & 0.908 & \textbf{0.170} & 0.950 & \textbf{4.290} & \textbf{0.240} & \underline{0.759} \\
MiniCPM-Duplex$^{\S}$ & 0.613 & 0.731 & 0.582 & 0.003 & 0.960 & 0.916 & 0.610 & 1.0 & 1.14 & 0.447 & 0.598 \\
\textcolor{gray}{Gemini Live$^{\dagger}$} & \textcolor{gray}{0.255} & \textcolor{gray}{0.310} & \textcolor{gray}{0.091} & \textcolor{gray}{0.012} & \textcolor{gray}{0.896} & \textcolor{gray}{0.655} & \textcolor{gray}{1.301} & \textcolor{gray}{0.891} & \textcolor{gray}{3.376} & \textcolor{gray}{1.183} & \textcolor{gray}{0.778} \\
\midrule
DuplexCascade & \textbf{0.058} & \textbf{0.222} & \textbf{0.218} & 0.009 & 0.949 & 0.832 & 1.724 & \underline{0.955} & \underline{4.016} & 1.225 & \textbf{0.858} \\
DuplexCascade-$\beta$ & \underline{0.343} & 0.458 & 0.309 & \underline{0.034} & \underline{0.811} & 0.899 & 0.567 & 0.950 & 4.011 & 0.850 & 0.748 \\
\bottomrule
\end{tabular}}
\label{table: fdb}
\end{table*}

\subsection{Overview}
Fig.~\ref{fig:pipeline} illustrates the overall proposed streaming pipeline. Unlike conventional cascaded systems, the proposed pipeline does not rely on VAD. Instead, user audio is continuously fed into a streaming ASR module to produce partial text outputs in real time. The partial text outputs are periodically, e.g. every 0.6 s, aggregated into a ``micro-turn" and sent to an LLM.

The proposed LLM is designed to handle a micro-turn sequence, where user and system micro-turns are interleaved using a dedicated end-of-turn token \texttt{<EOS>}. Given the micro-turn history, the LLM predicts the next system micro-turn and then sends it to a streaming TTS module, which synthesizes system audio incrementally. Importantly, the micro-turn does not necessarily contain speech content, but can contain special tokens that control  the conversational flow.

\subsection{Conversational Special Tokens}
To make turn-taking decisions controllable under streaming constraints, we introduce conversational special tokens.

\subsubsection{User's special token.}
\begin{itemize}
    \item \texttt{<no voice>}: insert when the buffer is flushed but no text has been recognized, to represent user silence for the current $\Delta t$ interval.
\end{itemize}

\subsubsection{System's special tokens.}

\begin{itemize}
    \item \texttt{<user is speaking>}: output when the user is still speaking, to represent the system should stay silent; the system micro-turn ends immediately with \texttt{<EOS>}.
    \item \texttt{<user finish speaking>}: output when the user has finished speaking, to represent the system should start responding; this token is followed by the system utterance content.
    \item \texttt{<user is interrupting>}: output when the user interrupts while the system is speaking, to represent the system should stop generation; the system micro-turn ends immediately with \texttt{<EOS>}.
    \item \texttt{<user backchannel>}: output when the user produces a backchannel while the system is speaking, to represent the system should ignore it and continue its current utterance.
    \item \texttt{<user is thinking>}: output when the user is silent after the system finishes response and is likely thinking, to represent the system should take no action; the system micro-turn ends immediately with \texttt{<EOS>}.
    \item \texttt{<system backchannel>}: output when the system should emit a short backchannel during user speech; once the LLM outputs \texttt{<system backchannel>}, we immediately play a pre-synthesised backchannel audio clip (sampled at random) and end the micro-turn with \texttt{<EOS>}.
\end{itemize}

\subsection{Dynamic Construction of Duplex Training Data}

\begin{figure}[h]
    \centering
    \includegraphics[width=0.48\textwidth]{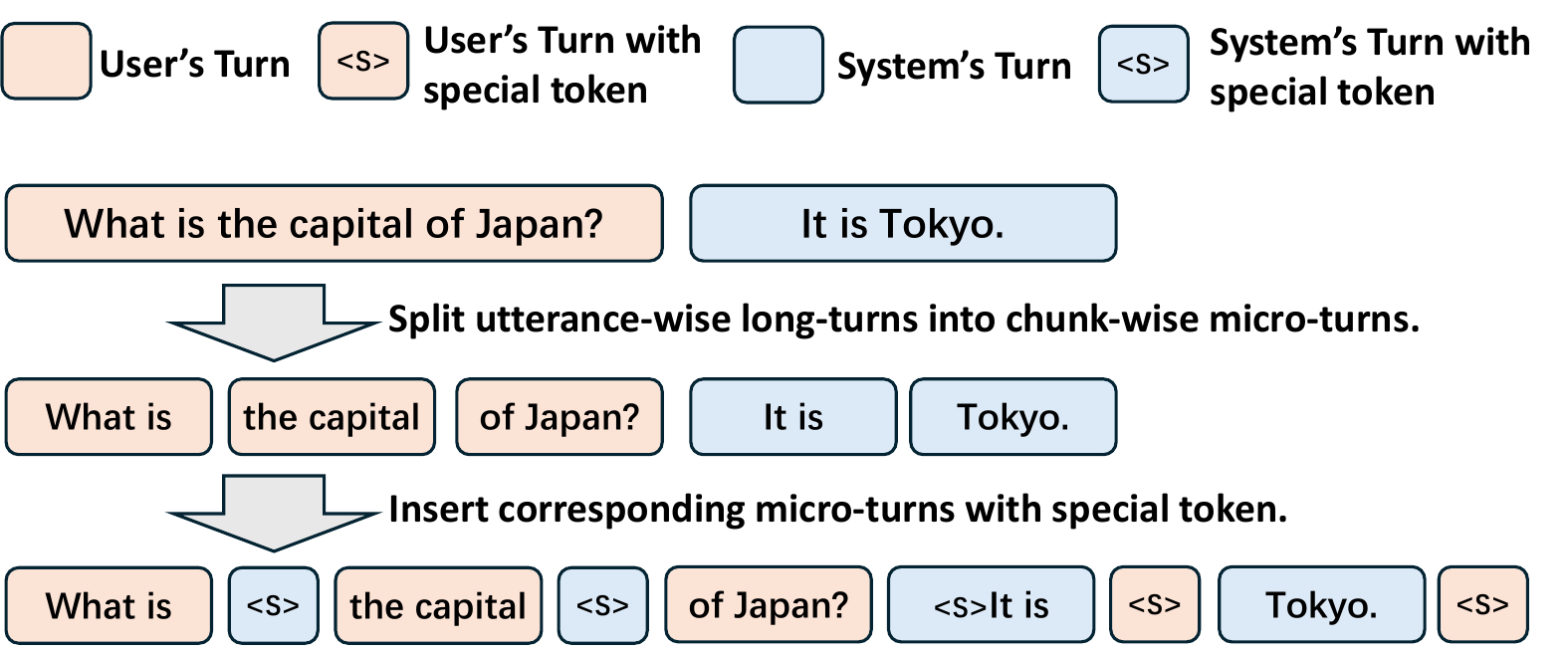}
    
    \caption{\normalfont Proposed dynamic construction pipeline for duplex training sequences from text-only dialogues.}
    \label{fig:data_construction}
\end{figure}

Real full-duplex dialogue corpora with turn-taking annotations are scarce. We therefore randomly sample 50k dialogues from UltraChat~\cite{ultrachat} and construct duplex-style training sequences dynamically by simulating 6 key interaction phenomena.

\subsubsection{Convert text dialogue into micro-turn segmentation}
As illustrated in Fig.~\ref{fig:data_construction}, we split both user and system utterance-wise long-turns into chunk-wise micro-turns. After each user micro-turn, we insert a system micro-turn that contains only \texttt{<user is speaking>} to indicate that the user continues. After each system micro-turn, we insert a user micro-turn \texttt{<no voice>} to represent that the user is silent. For the first system micro-turn of each original system turn, we prepend \texttt{<user finish speaking>} before the content to prompt the system to take the turn. During training, we apply next-token prediction LoRA fine-tuning~\cite{hu2022lora} only on system micro-turns, which helps preserve the backbone LLM's conversational intelligence.

\subsubsection{Simulating duplex interaction phenomena}
To better approximate real dialogue conditions, we simulate the following interaction phenomena.

\noindent\textbf{Randomized micro-turn length.} To emulate the variability of streaming ASR text emission, we sample the user micro-turn length uniformly from 1--7 tokens randomly. For training stability, we fix the system micro-turn length to 10 tokens.

\noindent\textbf{Natural pauses.} After each user micro-turn that are split from the original long turn, with probability 0.10 we randomly insert 1--5 additional silent user micro-turns represented by \texttt{<no voice>}. For these inserted silent micro-turns, the corresponding system micro-turns are supervised to output \texttt{<user is speaking>}, indicating that the system should refrain from taking the turn.

\noindent\textbf{User interruption.} For each system long turn, with probability 0.30 we simulate an interruption by letting the user start the next question at a randomly selected system micro-turn boundary. Upon receiving the first interrupting user micro-turn, the system is supervised to output \texttt{<user is interrupting>} and abandon the remainder of the interrupted response; during the subsequent user continuation, system micro-turns are supervised to output \texttt{<user is speaking>}.

\noindent\textbf{User backchannels.} While the system is speaking, each system micro-turn independently has a probability of 0.01 of replacing the next user \texttt{<no voice>} with a short textual backchannel (e.g., ``yes'', ``okay''). In this case, the system is supervised to output \texttt{<user backchannel>} and continue generating the ongoing response.

\noindent\textbf{System backchannels.} Because text-only dialogues rarely contain explicit system backchannels, we post-process user turns with Qwen2-72B-Instruct\footnote{\scriptsize \url{https://huggingface.co/Qwen/Qwen2-72B-Instruct}} to insert marker \texttt{<BC/>} after sentence-level units where a backchannel is appropriate. During training, these markers are used to supervise the system to output the token \texttt{<system backchannel>}.

\noindent\textbf{User thinking.} After the system finishes its response, we randomly insert 1--20 additional silent user micro-turns represented by \texttt{<no voice>}. For these inserted silent micro-turns, the corresponding system micro-turns are supervised to output \texttt{<user is thinking>}, indicating that the system should wait while the user is processing the response.


\section{Experiments}

\begin{table*}[t]
\centering
\caption{\normalfont VoiceBench results. The best-performing duplex dialogue model is shown in \textbf{bold}, and the second best is \underline{underlined}. $^{\S}$ means the results are infered from offical checkpoints. $^{\dagger}$ means the results directly obtained
from the paper.}
\label{tab:voicebench_selected}
\renewcommand{\arraystretch}{1.15}
\resizebox{\textwidth}{!}{
\begin{tabular}{l c c c c c c c c c c}
\hline
\textbf{Model} & \textbf{AlpacaEval} & \textbf{CommonEval} & \textbf{WildVoice} & \textbf{SD-QA} & \textbf{MMSU} & \textbf{OBQA} & \textbf{BBH} & \textbf{IFEval} & \textbf{AdvBench} & \textbf{Overall} \\
\hline
\makecell[l]{DSM-ASR+\\ Qwen2-7B-Instruct} & 4.57 & 3.87 & 3.73 & 46.42 & 58.52 & 72.09 & 58.10 & 51.29 & 97.12 & 69.66 \\
\midrule
Freeze-Omni$^{\dagger}$ & 4.03 & 3.46 & 3.15 & \textbf{53.45} & 28.14 & 30.98 & 50.70 & 23.40 & 97.30 & 55.20 \\
Moshi$^{\dagger}$ & 2.01 & 1.60 & 1.30 & 15.64 & 24.04 & 25.93 & 47.40 & 10.12 & 44.23 & 29.51 \\
PersonaPlex$^{\S}$ & 2.69 & 2.25 & 1.95 & 18.77 & 24.92 & 24.40 & 49.40 & 11.95 & 8.07 & 30.59 \\
\midrule
DuplexCascade & \textbf{4.40} & \underline{3.64} & \textbf{3.56} & 45.56 & \textbf{52.86} & \underline{56.04} & \textbf{59.80} & \underline{43.38} & \textbf{99.04} & \underline{65.41} \\
DuplexCascade-$\beta$ & \underline{4.38} & \textbf{3.70} & \underline{3.53} & \underline{46.35} & \underline{52.84} & \textbf{58.46} & \underline{59.00} & \textbf{44.45} & \underline{99.03} & \textbf{65.81} \\

\hline
\end{tabular}
}
\label{table:voicebench}
\end{table*}

\subsection{Implementation Details}
\label{sec:impl}
We implement the proposed cascaded pipeline with streaming ASR and streaming TTS based on DSM-ASR and DSM-TTS~\cite{kyutai2025streaming}. Following Freeze-Omni, we adopt Qwen2-7B-Instruct\footnote{\scriptsize \url{https://huggingface.co/Qwen/Qwen2-7B-Instruct}} as the text LLM backbone and perform lightweight adaptation on our dynamically constructed duplex training set using LoRA~\cite{hu2022lora} on the query and value projections (LoRA rank $r{=}16$, $\alpha{=}32$). The embedding parameters of the conversational special tokens are initialized from a zero-mean Gaussian distribution.

During adaptation, we fully fine-tune the original token embedding matrix, the embeddings of the newly introduced special tokens, and the token prediction head, while the remaining backbone parameters are updated through LoRA. To address class imbalance among conversational special tokens, we use a weighted loss with the following weights: \texttt{<user is speaking>} $=1$, \texttt{<user finish talking>} $=10$, \texttt{<user is interrupting>} $=5$, \texttt{<user backchannel>} $=2$, \texttt{<user is thinking>} $=1$, and \texttt{<system backchannel>} $=3$.

We train two variants: DuplexCascade (without system backchannels) and DuplexCascade-$\beta$ (with system backchannels), to evaluate whether training LLM with \texttt{<system backchannel>} affects overall model performance. DuplexCascade does not include the \texttt{<system backchannel>} token and is trained without any system-backchannel supervision, whereas DuplexCascade-$\beta$ enables this token and is trained with the corresponding simulation described in Sec.~3.3.

Optimization uses AdamW~\cite{loshchilov2017decoupled} with learning rate $1\times 10^{-5}$ and a linear warmup schedule for 500 steps. We set the batch size to 32, fine-tune for 5k steps, and use a maximum sequence length of 4096 tokens. Each model was trained using 8 NVIDIA H100 GPUs for only 5 hours.

\subsection{Full-Duplex-Bench Results}
\label{sec:fdb}
We evaluate DuplexCascade on Full-Duplex-Bench~\cite{lin2025full} against other state-of-the-art
models. The benchmark reports Take-Over Rate (TOR) for each condition; however, TOR has mixed directions across subsets, making it inconvenient to summarize overall turn-taking quality.

To provide an aggregated view, we introduce \emph{Averaged Turn-Taking Accuracy}. For subsets where lower TOR is better (Pause Handling and Backchannel), we define accuracy as $1{-}\mathrm{TOR}$. For subsets where higher TOR is better (Smooth Turn Taking and User Interruption), we define accuracy as $\mathrm{TOR}$. Averaged Turn-Taking Accuracy is the unweighted mean of these accuracies across all TOR-reported subsets.

Table~\ref{table: fdb} shows that DuplexCascade achieves the best Averaged Turn-Taking Accuracy among the evaluated open systems, indicating consistently strong turn-taking decisions across diverse duplex scenarios. In contrast, Freeze-Omni, which uses VAD-driven approaches that rely on an external endpointing head, exhibits noticeably weaker turn-taking robustness, highlighting the benefit of explicitly modeling duplex control decisions via LLM tokens rather than inferring them from external endpointing head.


And DuplexCascade-$\beta$ achieved the second-best results on backchannel-related metrics (ICC Freq and JSD), while competitive on overall turn-taking accuracy. The results suggest that our proposed method can control system's reaction styles by text-only LLM training. 

\subsection{VoiceBench Results}
\label{sec:voicebench}

A practical spoken dialogue system must also preserve the reasoning and instruction-following capability of the underlying text LLM. We therefore evaluate conversational intelligence using VoiceBench~\cite{chen2024voicebench}. In addition to other state-of-the-art duplex models, to assess how our method affects the LLM's conversational intelligence, we also implement a naive pipeline as defined in VoiceBench~\cite{chen2024voicebench}, which combines DSM-ASR~\cite{kyutai2025streaming} and Qwen2-7B-Instruct.

As shown in Table~\ref{table:voicebench}, our models substantially outperform prior duplex systems across nearly all VoiceBench dimensions, suggesting that LoRA fine-tuning with text (instead of speech) helps preserve conversational intelligence by avoiding cross-modality alignment issues. Compared with the naive DSM-ASR+Qwen2-7B-Instuct pipeline, DuplexCascade and DuplexCascade-$\beta$ achieve competitive scores with a moderate gap, suggesting that our text-only adaptation largely retains the backbone LLM's capabilities despite operating under streaming micro-turn constraints. Notably, DuplexCascade-$\beta$ demonstrates that our method enables the LLM to produce backchannels while preserving its core capabilities.


\subsection{\texorpdfstring{Micro-Turn $\Delta t$ Analysis}{Micro-Turn Delta t Analysis}}
\begin{figure}[t]
    \centering
    \includegraphics[width=0.5\textwidth]{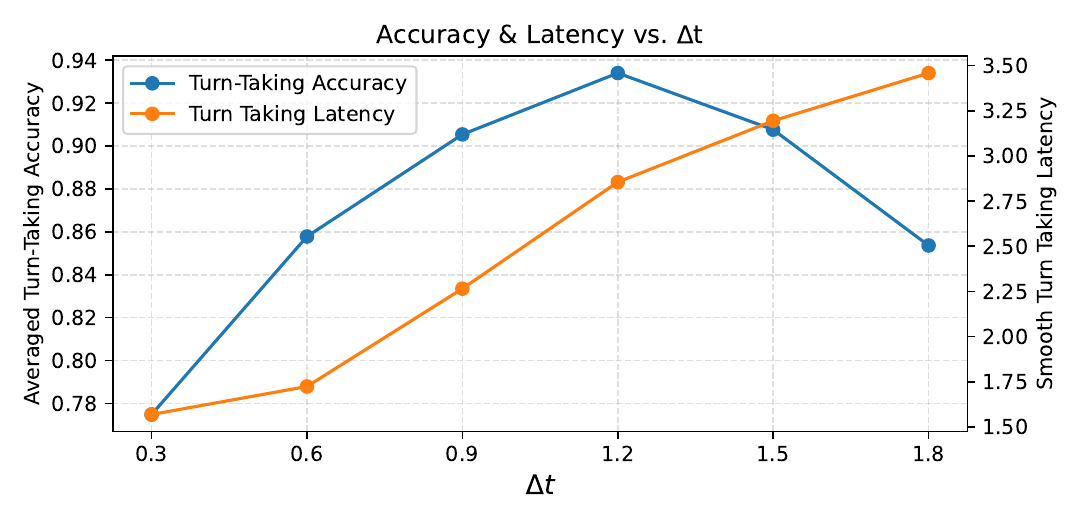}
    \caption{\normalfont Effect of the micro-turn duration $\Delta t$ on Averaged Turn-Taking Accuracy and turn-taking latency on Full-Duplex-Bench.}
    \label{fig:ablation}
\end{figure}

We study how the micro-turn duration $\Delta t$ affects turn-taking accuracy and response latency. Specifically, we evaluate $\Delta t \in \{0.3, 0.6, 0.9, 1.2, 1.5, 1.8\}\,$s on Full-Duplex-Bench, and report the resulting Averaged Turn-Taking Accuracy and Smooth Turn-Taking Latency in Fig.~\ref{fig:ablation}. We observe that Averaged Turn-Taking Accuracy improves as $\Delta t$ increases up to $1.2\,$s, after which it degrades. In contrast, Smooth Turn-Taking Latency increases monotonically with larger $\Delta t$, since the system must wait longer before each buffer flush. These results suggest that, under our simulation setting, $\Delta t{=}1.2\,$s provides the strongest turn-taking performance, but at the cost of higher latency. We therefore choose $\Delta t{=}0.6\,$s as a practical trade-off between turn-taking accuracy and latency.

\section{Conclusion}
In this work, we presented DuplexCascade, a practical cascaded pipeline for full-duplex speech-to-speech dialogue. By adapting a strong text LLM with only a small amount of text-only dialogues via lightweight LoRA fine-tuning, DuplexCascade achieves VAD-free and robust turn-taking ability while largely preserving the conversational intelligence of the underlying LLM. These results highlight that full-duplex interaction can be enabled within a modular cascade without sacrificing the strengths of modern text LLMs.





\section{Generative AI Use Disclosure}
Generative AI tools were used for language editing and improving the phrasing of this manuscript.

\bibliographystyle{IEEEtran}
\bibliography{mybib}

\end{document}